\documentclass[runningheads]{llncs}
\usepackage[T1]{fontenc}
\usepackage{newtxtext}       %
\usepackage{bbding}
\usepackage[varvw]{newtxmath}       
\usepackage{xspace}
\newcommand{\ours}{\textsf{PAI-KNN}\xspace}

\usepackage{graphicx}
\begin{document}
\title{Inference-Time Prior Adaptation for Imbalanced Data Streams under Concept Drift}
%
%
\author{Tobias Callies\inst{1}\orcidID{0009-0004-5852-4103}\Envelope \and
Eirini Ntoutsi\inst{1}\orcidID{0000-0001-5729-1003} \and
Arthur Zimek\inst{2}\orcidID{0000-0001-7713-4208}}
\authorrunning{T. Callies et al.}
%
\institute{University of the Bundeswehr Munich, Munich, Germany \\
\email{\{tobias.callies,eirini.ntoutsi\}@unibw.de}
 \and
University of Southern Denmark, Odense, Denmark\\
\email{zimek@imada.sdu.dk}}
\maketitle              
\begin{abstract}
Data stream classification under concept drift and class imbalance requires models to adapt continuously while maintaining strong minority-class performance. Many existing approaches rely on \emph{active memory management}, including instance removal, memory restructuring, resampling, and synthetic data generation.
We present \ours{}, a $k$-nearest neighbor classifier that adapts via inference-time prior adjustment rather than memory modification. \ours{} maintains a simple sliding-window buffer and dynamically adjusts its decision rule using class prior estimates derived from recent observations.
Evaluated on a multitude of synthetic data streams covering various imbalance and drift scenarios, \ours{} achieves competitive performance with state-of-the-art methods, performing particularly well under scenarios involving dynamic class imbalance and concept drift. These findings highlight the potential of inference-time adaptation in challenging non-stationary environments and suggest that adaptation in memory-based stream classifiers need not rely exclusively on memory modification.

\end{abstract}
\section{Introduction}
\label{sec:intro}
Data stream classification~\cite{read2025supervised} often requires models to operate under two simultaneous challenges: \emph{concept drift} and \emph{class imbalance}. As data distributions evolve over time, classifiers must continuously adapt to changing data while maintaining reliable performance on underrepresented classes. This problem arises in numerous real-world applications, including fraud detection, network monitoring, and predictive maintenance.
Existing approaches address concept drift and class imbalance in different ways. Many stream classifiers are \emph{model-based}, maintaining either a single evolving model~\cite{hulten2001mining}, \cite{wagner2015ageing} or a pool of models that are continuously updated, reweighted, or replaced over time ~\cite{cano_rose_2022}, \cite{loezer_cost-sensitive_2020},~\cite{gomes_adaptive_2017}, \cite{wang_dealing_2016}, often combined with drift detectors like ADWIN~\cite{bifet_learning_2007}.
$K$-nearest neighbor (KNN) classifiers, in contrast, are \emph{instance-based or lazy learners} that maintain explicit memories of previously observed instances and perform prediction directly from stored observations.
Several KNN-based stream classifiers have been proposed that address concept drift and class imbalance through increasingly sophisticated \emph{ active memory management mechanisms} \cite{losing_knn_2016,abolfazli_drift-aware_2020}. Adaptation is typically achieved through instance removal, restructuring, compression, or balancing operations. While effective, active buffer-management strategies increase algorithmic complexity and may prematurely alter or discard useful information.
In this paper, we investigate whether adaptation in KNN-based stream classifiers must occur through memory modification. We propose \ours{}, a $k$-nearest neighbor classifier that maintains a simple sliding-window memory and adapts through inference-time prior adjustment. By shifting adaptation from memory management to inference, \ours{} responds to changing class distributions without the need for expensive memory management mechanisms. 
We evaluate \ours{} on a benchmark of synthetic data streams~\cite{aguiar_survey_2023} covering diverse imbalance and drift scenarios. Despite its simplicity, \ours{} achieves competitive performance with state-of-the-art methods and performs particularly well under scenarios involving high class imbalance and concept drift. These findings suggest that inference-time prior adaptation can provide a simple and effective alternative to complex memory management by avoiding structural
assumptions.
The rest of the paper is organized as follows: Section~\ref{sec:basics} introduces the preliminaries and basic concepts. Related work is discussed in Section~\ref{sec:related}. In Section~\ref{sec:oour}, we present our proposed method~\ours. The experimental evaluation is detailed in Section~\ref{sec:experiments}. Conclusions and outlook are provided in Section~\ref{sec:conclusions}.

\section{Preliminaries and basic concepts}
\label{sec:basics}
A data stream is a potentially infinite sequence  of instances $D=(x_1, \dots, x_t, \dots)$ where $x_t\in \mathbb{R}^d$ denotes 
a $d$-dimensional feature vector arriving at time $t$.
Under the standard prequential evaluation protocol,
an observation $x_t$ is first classified, producing a prediction $\hat{y}_t$. The true label $y_t$ is revealed subsequently and can be
used to update the classifier. Instance-label pairs $(x_t,y_t)$ are assumed to be sampled from evolving data distributions $\mathbb{P}_t(x,y)$.
Concept drift~\cite{gama_survey_2014} may hence arise from \emph{class priors} (${\mathbb{P}_{t+1}(y)\neq\mathbb{P}_t(y)}$), \emph{likelihoods} (${\mathbb{P}_{t+1}(x|y)\neq\mathbb{P}_t(x|y)}$), or  \emph{posteriors} (${\mathbb{P}_{t+1}(y|x)\neq\mathbb{P}_t(y|x)}$) changing.

For classification problems, the \emph{imbalance ratio} $\mathit{IR}$ quantifies the class-prior distribution and is defined as the (potentially infinite) ratio of majority to minority class frequencies over a set of instance labels (in case of multi-class problems, ratio of largest to smallest class).
Consequently, changes in class priors are directly reflected in changes in the imbalance ratio. Since \ours{} adapts through dynamic prior adjustment, such changes are of particular interest in this work.
In many real-world streams, class frequencies do not only exhibit strong imbalance but may also change over time, resulting in \emph{dynamic imbalance ratios}. Depending on the speed of change, such shifts may occur either \emph{suddenly} or \emph{gradually}.
Due to the potentially infinite nature of data streams and processing speed requirements, memory-based classifiers typically maintain only a bounded subset of previously observed instances (also known as \emph{buffer}).

\section{Related work}
\label{sec:related}
We divide related work into three parts: classifiers for imbalanced data streams (Section~\ref{sec:relatedImbalancedStreams}), KNN classifiers for data streams (Section~\ref{sec:relatedKNNs}) and KNN classifiers for imbalanced static data (Section~\ref{sec:relatedStaticKNN}).
\subsection{Imbalanced stream classifiers}
\label{sec:relatedImbalancedStreams}
Tree-based methods and their ensembles remain the dominant approach to stream classification. For imbalanced streams, numerous methods have been proposed that combine drift detection and adaptation with resampling, cost-sensitive learning, or imbalance-aware ensemble construction \cite{aguiar_survey_2023}. For the purpose of this work, the distinction between inference-based interventions (adjustments to the prediction rule) and data-based interventions (modifications of the data presented to the learner) is particularly important.

We therefore restrict our comparison for such methods to two representative state-of-the-art methods selected due to their performance~\cite{aguiar_survey_2023}: 
\emph{Cost-sensitive Adaptive Random Forest} (CSARF) \cite{loezer_cost-sensitive_2020} introduces a cost matrix to tackle imbalanced performance using adaptive random forests at inference time. 
\emph{Synthetic Minority Oversampling Technique with Online Bagging} (SMOTE-OB) \cite{bernardo_smote-ob_2021} is the best-performing method among methods heavily influenced by the SMOTE \cite{chawla_smote_2002} algorithm, generating synthetic data on subsets of features, which are then used to continually update an ensemble of tree learners. 

\subsection{KNN-based stream classifiers }
\label{sec:relatedKNNs}

Self-Adjusting Memory KNN (SAM-KNN) \cite{losing_knn_2016} addresses concept drift through active memory management. The method maintains a short-term memory (STM) containing recent observations and a long-term memory (LTM) storing compressed historical information. As the stream evolves, SAM-KNN continuously restructures these memories based on the label consistency of stored instances with recent observations. Following drift detection, instances that remain consistent with the STM are retained, whereas inconsistent instances may be permanently removed. Adaptation is therefore achieved through active modifications of the stored memory.

Drift-Aware Multi-Memory Model (DAM3) \cite{abolfazli_drift-aware_2020} extends SAM-KNN to handle class-imbalanced streams. To improve minority-class performance, DAM3 introduces a third Working Memory (WM) and explicitly maintains a balanced class representation while identifying and removing class inconsistencies in the stored data. As in SAM-KNN, adaptation is achieved through active management of the underlying memories.

Both approaches achieve adaptation through active buffer management, i.e., modifications of the stored memories. SAM-KNN and DAM3 continuously inspect, restructure, and curate stored instances to maintain memory consistency and adapt to changing stream characteristics. While effective, such mechanisms require ongoing memory interventions and may permanently alter or discard information based on potentially uncertain evidence of change. 
This is particularly consequential in evolving streams, where instances initially regarded as outliers or noise may later prove representative of a changing class distribution or decision boundary~\cite{Masud2011NovelClasses}. Prematurely discarding them may therefore remove valuable evidence of concept evolution.

As our proposed adjustment is independent of the metric, we consider metric-learning approaches \cite{kummert_metric_2023,mahalanobis} and extensions as orthogonal to this work's focus point.

\subsection{Static inference-time prior-adjusted KNN}
\label{sec:relatedStaticKNN}
An inference-time prior-adjusted KNN for static imbalanced learning was proposed in~\cite{gottcke_handling_nodate}. 
The method adjusts the decision rule rather than the data to compensate for class imbalance and achieves performance comparable to data-level interventions such as SMOTE~\cite{chawla_smote_2002} at a substantially lower 
computational cost.

In this formulation, KNN classification is viewed as a maximum-a-posteriori (MAP) estimator, where the class priors $\mathbb{P}(y_i)=N_i/N$ are estimated from the data and the likelihood is approximated by $\mathbb{P}(x|y_i)=\frac{n_i}{N_i}V(x)$. Here  $n_i$ is the number of instances of class $y_i$ among the $k$ nearest neighbors of an instance $x$, $N_i$ is the total number of instances of class $y_i$ in the dataset, $N$ is the total number of instances, $m$ will denote the number of classes, and $V(x)$ is the volume of the neighborhood. 

Under mild assumptions, and using Bayes' theorem, the posterior can then be written as:  
\begin{equation}
\mathbb{P}(y_i \mid x) \propto \frac{\frac{n_i}{N_i}V(x)\frac{N_i}{N}}{\sum_{j=1}^m \frac{n_j}{N_j}V(x)\frac{N_j}{N}} = \frac{n_i}{\sum_{j=1}^mn_j} = \frac{n_i}{k}     
\end{equation}

To improve performance on underrepresented classes, the prior estimate is substituted by a uniform prior: $\mathbb{P}(y_i)=\frac{1}{m}$. The new decision rule then becomes: $\hat{y} = \arg\max_{\{j=1,\dots,m\}} \frac{n_j}{N_j}$.

In this work, we extend such inference-time prior adjustment from static imbalanced learning to non-stationary data streams with evolving class priors.

\section{\ours{}: Inference-Time Prior Adaptation KNN}
\label{sec:oour}
Motivated by the promising results of the prior-adjusted static KNN~\cite{gottcke_handling_nodate} for static imbalance learning and by the prevalence of increasingly sophisticated active memory management mechanisms in KNN-based stream classifiers, we investigate 
whether adaptation can be achieved through inference-time prior adjustment rather than active memory management. To this end, we extend the concept of inference-time prior adjustment to non-stationary data streams.

\begin{figure}[tbp]
\centerline{\includegraphics[width=0.8\columnwidth]{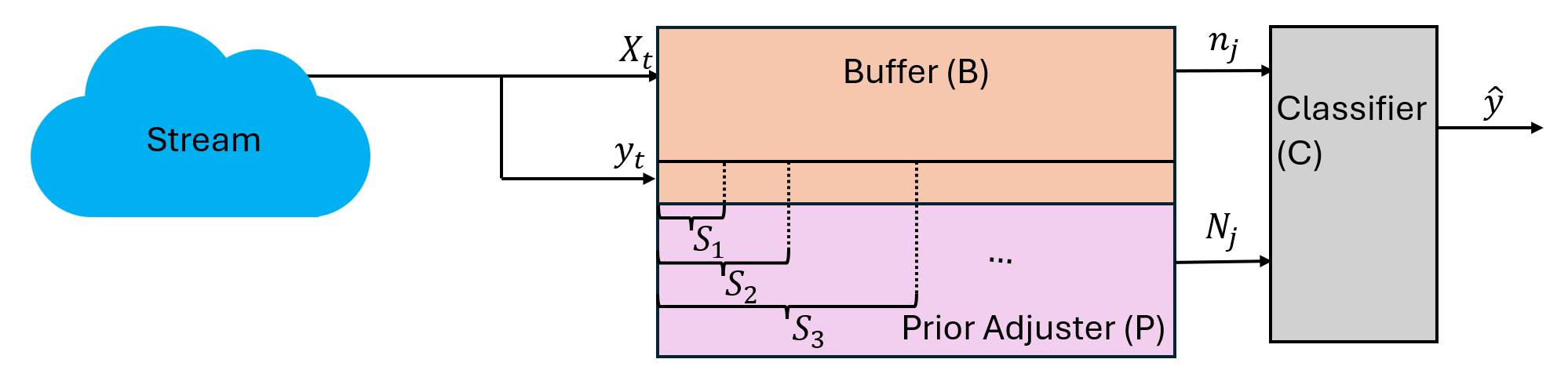}}
\Description{Schematic Model Overview presenting PAI-KNN}
\caption{Model components (B),(P),(C). Target labels are received after prediction}
\label{fig:architecture}
\end{figure}

An overview of the proposed approach is shown in Fig.~\ref{fig:architecture}. Relative to a vanilla KNN classifier, \ours{} augments prediction with a prior-adjustment mechanism that adapts the decision rule to changing class distributions over time. Section~\ref{sec:ourModel} introduces the model components, followed by the update and inference steps in Section~\ref{sec:ourLearn}. Computational complexity is discussed in Section~\ref{sec:ourCompComplexity}. 

\subsection{Model definition}
\label{sec:ourModel}
\ours augments a standard sliding-window KNN classifier with a prior-adjustment module that estimates class priors from recent observations and incorporates them into the prediction rule. Concretely, \ours consists of three components explained hereafter: 
\begin{itemize}
    \item a \textbf{buffer} $B$ of size $N$
    \item a \textbf{prior-adjustment module} $P$, which estimates class priors from recent stream history
    \item a \textbf{KNN classifier} $C$, which combines neighborhood information with the estimated priors during inference.
\end{itemize}

\noindent \textbf{Buffer $B$:} The buffer $B$ is implemented as a FIFO queue and stores recent observations from the stream.

\noindent \textbf{Prior-adjuster $P$:} 
The prior-adjustment module $P$ estimates the current class prior distribution from recent stream observations. To do so, \ours{} maintains multiple sliding windows of different sizes. The multiple windows allow the classifier to adaptively select the longest statistically consistent history for prior estimation, thereby limiting responsiveness to statistically significant changes. Class prior estimates are then obtained from the label frequencies within the selected window and incorporated into the prediction rule.

Concretely, we use $L$ sliding windows $S_1,\dots,S_L$ that store the latest \break $2^{l_{\min}},\cdots,2^{l_{\max}}$ labels, which can be effectively implemented using the existing buffer queue.  
The largest window is chosen to match the buffer size, i.e., $l_{\max}=\log_2(|B|)$,
and $l_{\min}\geq5$ is recommended for efficient hypothesis testing. 
At time $t$, the sliding windows are given by 
\begin{equation}
S_k(t) = \{y_i \mid t-i\leq 2^{l_{\min}+k-1}, i\geq 1\}    
\end{equation}

The module further maintains a \emph{significance threshold} $p$ used to compare label distributions across windows.
During inference, a $\chi^2$ test is used to assess whether two windows are plausibly generated from the same distribution.

\noindent \textbf{The KNN classifier $C$:} 
The KNN classifier $C$ operates on the buffer $B$ and is associated with the typical KNN hyperparameters: the number of neighbors $k$ and the distance function $d(\mathbb{R}^d\times \mathbb{R}^d)\rightarrow \mathbb{R}^+$.
In addition, 
$C$ includes a parameter $\theta\in[0, 1]$, interpolating between uniform priors and the observed class priors.
This interpolation provides a direct trade-off between overall accuracy and minority-class performance, with $\theta=1$ corresponding to classical KNN and $\theta=0$ optimizing for recall on the minority class, as in knn-BPP \cite{gottcke_handling_nodate}. 

\subsection{Model update and inference}
\label{sec:ourLearn}
\noindent \textbf{Update step:} it only consists of queue insertions of $(x, y)$ into the buffer $B$, and an insertion of the label $y$ into the sliding windows $S_j$. 

\noindent \textbf{Inference step:} For an instance $x$, the classifier $C$ determines for every class $y_j\in Y$ the number of instances $n_j$  among the $k$ nearest neighbors of $x$. 
The prior-adjustment module $P$ selects the longest statistically consistent history and uses the corresponding label frequencies to estimate the current class priors.
In particular, $P$ provides for every class the number of samples  $N_j=|\{y = y_j \mid y\in S^*\}|$ in the longest consistent sliding window $S^*=S_{k^*}$, where 
\begin{equation}
k^*=\min\left( \{1\leq k < L\mid \chi^2(S_k, S_{k+1}\setminus S_k)< p\}\cup\{L\}\right)
\end{equation}
With all $n_j$ and $N_j$ determined that way, $C$ adjusts the prior using $\theta$ to effectively choose the sensitivity to imbalance and calculates the following posterior: 
\begin{equation}
p(y_i\mid x) = \frac{\frac{n_i}{N_i}\left(\theta\frac{N_i}{N}+(1-\theta)\frac{1}{m}\right)}{\sum_{j=1}^{|Y|}\frac{n_j}{N_j}\left(\theta\frac{N_j}{N}+(1-\theta)\frac{1}{m}\right)}    
\end{equation}
The class label of $x$ is the one that maximizes the posterior. 

\subsection{Computational Complexity}
\label{sec:ourCompComplexity}

The time complexity of the \emph{update step} is effectively negligible as only a queue insertion takes place. 
Time complexity in the \emph{inference step} is dominated by the neighborhood search and hence worst-case complexity naively scales with $\mathcal{O}(Nd^2)$, where $N$ is the buffer size and $d$ is the feature space dimensionality.

Compared to active buffer management methods, worst-case complexity is therefore significantly faster. For example, SAM-KNN scales slightly worse than $\mathcal{O}(N^2d^2)$ \cite{losing_knn_2016}. The space complexity of our method scales with $\mathcal{O}(Nd)$. 

\section{Experiments}
\label{sec:experiments}
We evaluate PAI-KNN on a benchmark of synthetic data streams against state-of-the-art competitors. Section~\ref{sec:expSetup} describes the evaluation design, datasets, competitors, metrics, and experimental protocol before presenting comparative results (Section~\ref{sec:ExpResults}) and a parameter analysis (Section~\ref{sec:ExpResultsParametersEffect}) with respect to the prior-adjustment trade-off parameter $\theta$, and the neighborhood size $k$.

\subsection{Evaluation design}
\label{sec:expSetup}

\noindent \textbf{Datasets:}

We adopt the synthetic stream benchmark~\cite{aguiar_survey_2023}. Streams consist of 200.000 instances, and are generated using multiple random generators to evaluate methods across different feature spaces.
We restrict analysis to (five) binary generators producing only numerical features to avoid the need to craft distance functions on categorical features.
Benchmark scenarios fall into four categories: i) static imbalance ratio, ii) dynamic imbalance ratio, iii) instance-level difficulties, and iv) concept drift. 

\noindent \textbf{Competitors:}
We compare our approach to the competitor methods discussed in Section~\ref{sec:related}: CSARF \cite{loezer_cost-sensitive_2020}, SMOTE-OB \cite{bernardo_smote-ob_2021}, and SAM-KNN ~\cite{losing_knn_2016}. For DAM3 (\cite{abolfazli_drift-aware_2020}), the available online implementations struggled with long stream sizes. 
For CSARF and SMOTE-OB, we use results and hyperparameters from a recent survey \cite{aguiar_survey_2023}, where they are set according to their self-reported standard configuration. 
For SAM-KNN, experiments were performed using the reference implementation in \cite{skmultiflow}. To validly compare against our own method, buffer size was set to 1024.\footnote{H-params SAM-KNN: $k=5$, $N=1024$, ltm\_size$=0.4$, min\_stm\_size$=50$, weighting$=$uniform, use\_ltm$=$True, stm\_size\_option$=$maxACCApprox} 
For PAI-KNN, we chose $N=1024$, sliding windows in the prior adjuster ranging from size $64$ to $1024$, $p=0.01$ as a statistical significance threshold and $k=20$. $\theta=0$ was chosen, thereby prioritizing G-Mean performance. 
Although the choice of distance function can have an important impact, it is orthogonal to this work. 
We therefore use the $L_2$ norm throughout to avoid confounding the effects of the prior intervention with those of the chosen metric. 

\noindent\textbf{Evaluation Setup:}
We run all experiments using River \cite{montiel2021river}, with data streams generated
using the code of ~\cite{aguiar_survey_2023}. In line with their work \cite{aguiar_survey_2023}, we report the median obtained from running methods on 5 streams generated with different random seeds. Evaluation metrics for each run are calculated by taking the arithmetic mean over sliding windows of size 500.
We aggregate metrics over groups of different synthetic streams, using the arithmetic mean. Code is available\footnote{\url{https://github.com/tcallies/streamKNN}}.

\noindent \textbf{Evaluation Metrics:}
Performance is evaluated using Accuracy, Cohen’s Kappa, and G-Mean. Accuracy measures overall correctness, Kappa accounts for chance agreement, and G-Mean emphasizes balanced performance across classes.
We do not report detailed computational costs as these are very platform and implementation dependent \cite{DBLP:journals/kais/KriegelSZ17}. To give some reference nonetheless, our not-optimized Python implementation of \ours processes approximately 1000 instances in two seconds, depending on the dimensionality of the feature space.

\subsection{Predictive performance}
\label{sec:ExpResults}
\begin{figure}[tbp]
\centerline{\includegraphics[width=0.9\columnwidth]{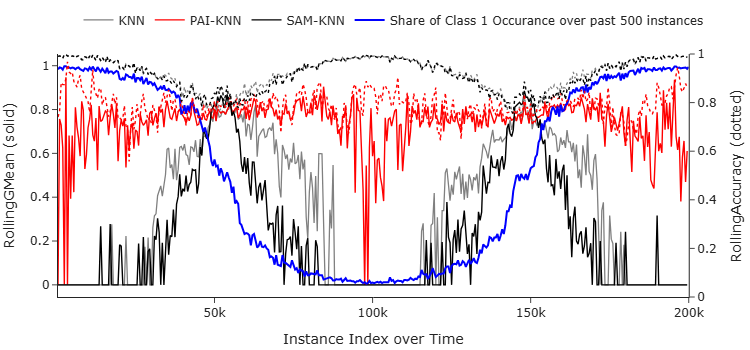}}
\Description{G-Mean and Accuracy plotted over time over time  -  HyperplaneGenerator, gradual reflipping}
\caption{G-Mean and Accuracy over time  -  HyperplaneGenerator, gradual reflipping}
\label{fig:dyn_time}
\end{figure}

To better understand the behavior of \ours{} under changing imbalance ratios, we plot in Fig.~\ref{fig:dyn_time} the performance for PAI-KNN, SAM-KNN, and a baseline KNN method (operating on a FIFO buffer without any prior adjustment, with $k=20$, $N=1024$) for a concrete scenario in which the imbalance ratio reverses twice over time (\emph{reflipping}). We see that SAM-KNN's imbalanced performance deteriorates notably more quickly than the baseline KNN method for high imbalance ratios. On the other hand, PAI-KNN maintains a high G-Mean level throughout even significant gradual changes. We interpret this as follows: maintaining buffer consistency is detrimental if streams change significantly and rapidly enough. 

\begin{figure}[tbp]
\centerline{\includegraphics[width=0.85\columnwidth]{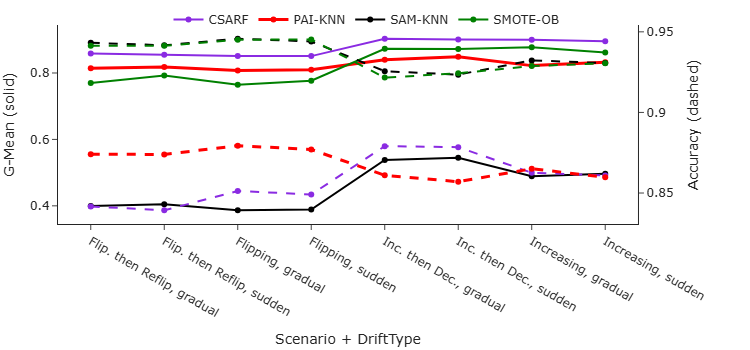}}
\Description{G-Mean and Accuracy per method for streams with dynamic IR}
\caption{G-Mean and Accuracy per method for streams with dynamic IR}
\label{fig:dyn_IR}
\end{figure}

This observation is reflected in the results shown in Fig.~\ref{fig:dyn_IR}, where we plot G-Mean and Accuracy for four different scenarios of imbalance ratio changes, with changes occurring either gradually or suddenly, and results aggregating over streams from different generators. \ours performs on par with ensemble competitors in terms of G-Mean, while underperforming in Accuracy.
Sudden changes describe an instantaneous and significant shift in the data generating distribution, which for gradual changes is prolonged through interpolation over several thousands timesteps. 

CSARF, similarly to us, excels in terms of G-Mean at the sacrifice of Accuracy, showcasing the natural conflict arising from cost-sensitive adjustments. 

\subsubsection{Concept Drift}
\begin{figure}[tbp]
\centerline{\includegraphics[width=0.85\columnwidth]{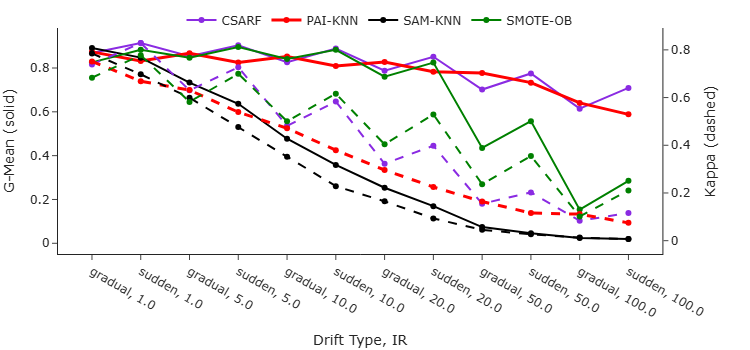}}
\Description{G-Mean and Kappa per method for concept drift streams}
\caption{G-Mean and Kappa per method for concept drift streams}
\label{fig:cd_static}
\end{figure}

We limit analysis to streams with both gradual and sudden feature space concept drift, for different, static imbalance ratios. We plot in Fig.~\ref{fig:cd_static} performance by drift type and imbalance ratio, and aggregated across generators. We see that in terms of G-Mean, we outperform competitors for gradual changes, and high imbalance ratios. Again, a very clear data-intrinsic tradeoff can be recognized that is resolved in favor of G-Mean for PAI-KNN and CSARF, whereas performance for SAM-KNN correlates more strongly with SMOTE-OB. 

\subsubsection{Static Imbalance Ratio}
\begin{figure}[tbp]
\centerline{\includegraphics[width=0.85\columnwidth]{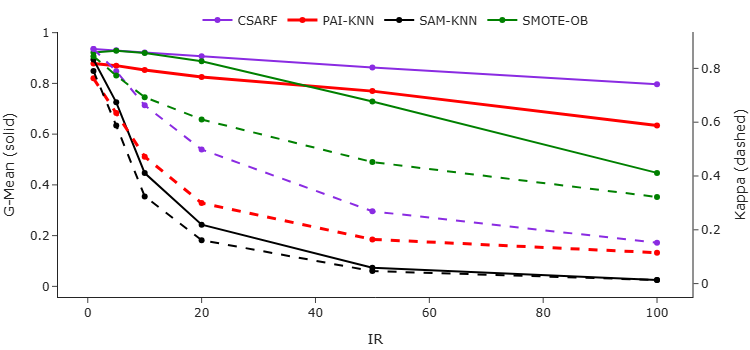}}
\Description{G-Mean and Kappa per method for static streams with varying imbalance ratio}
\caption{G-Mean and Kappa per method for static streams with varying $\mathit{IR}$}
\label{fig:static_IR}
\end{figure}

Streams in this category have no intrinsic temporal component; both feature and label distribution remain fixed over time. 
Here, the ability to retain and add information over time drives performance, whereas the ability to forget---a focus point of \ours---is irrelevant. 
As a result, \ours performs generally worse than competitors, but robust to very high $\mathit{IR}$ (Fig.~\ref{fig:static_IR}). 

\subsubsection{Instance Level Difficulties}
\begin{figure}[tbp]
\centerline{\includegraphics[width=0.85\columnwidth]{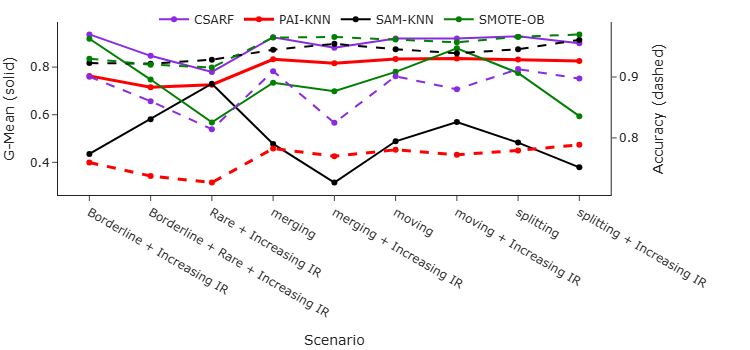}}
\Description{G-Mean and Accuracy per method for ILD streams with concept drift}
\caption{G-Mean and Accuracy per method for ILD streams with concept drift}
\label{fig:ild}
\end{figure}

Streams with instance level difficulties (ILD) incorporate one or multiple challenges to differentiate data: Imbalance ratios changes, class-specific concept drift, many instances in borderline regions or in rare locations. For those ILD streams involving concept drift we plot results in Fig.~\ref{fig:ild}.

In terms of G-Mean, we keep up with state of the art competitors in this case despite the generator synthesizing the ILD streams producing distributions not particularly well-suited to a $L_2$ distance function. 

We observe that SAM-KNN performs very well for the scenario involving rare instances. This is in line with our understanding that the buffer cleaning and management mechanism can perform very well when underlying assumptions are met and retaining information is important. PAI-KNN on the other hand performs more evenly across ILD scenarios: due to imposing only weak assumptions, we remain agnostic to different setups.  

\subsection{Parameter Effect}
\label{sec:ExpResultsParametersEffect}
\begin{figure}[tbp]
\centerline{\includegraphics[width=0.9\columnwidth]{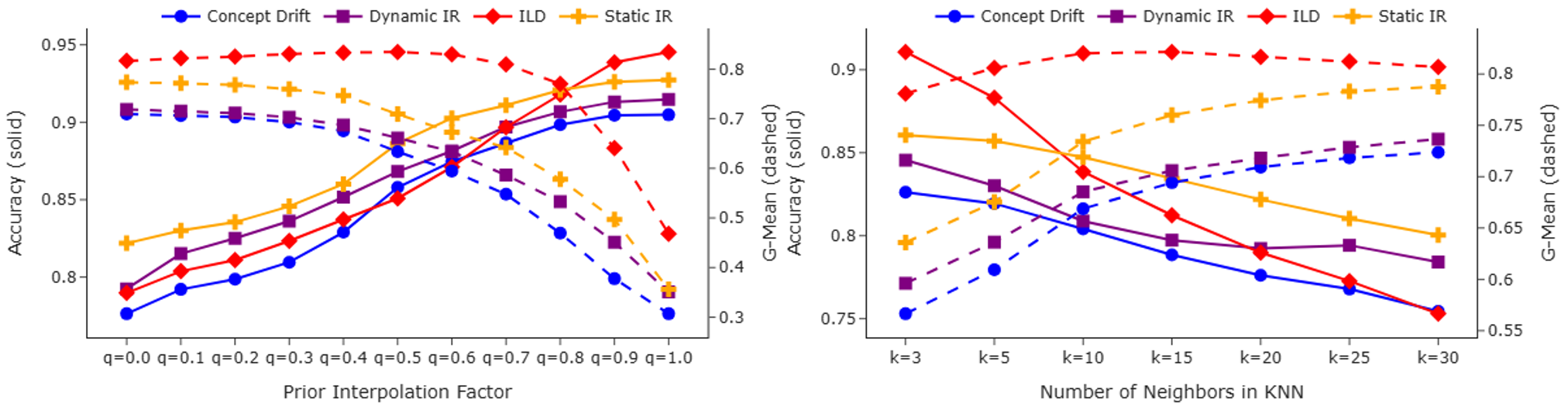}}
\Description{Effects of $\theta$ and $k$ per Experiment Group}
\caption{Effects of $\theta$ and $k$ per Experiment Group}
\label{fig:sweep_b2}
\end{figure}
\label{sec:EvalParamEffect}
We evaluate parameter effects using a subset of 36 streams, grouped by scenario class. 

For the \emph{trade-off parameter $\theta$}, $\theta=1$ maximizes accuracy, while $\theta=0$ optimizes G-Mean. Notably, G-Mean performance declines only slightly up to $\theta=0.4$, offering a beneficial tradeoff. For streams exhibiting instance level difficulties (ILD), G-Mean even slightly improves up to $\theta=0.5$, which showcases that under the right conditions, significant accuracy gains can be achieved using \ours without sacrificing on imbalanced performance.
For the \emph{neighborhood size $k$}, we again fix $\theta=0$. Low values of $k$ show higher accuracy while G-Mean benefits from larger $k$. 
Again, the optimal tradeoff is both dependent on performance preferences and stream characteristics.

\section{Conclusion}
\label{sec:conclusions}

We presented \ours{}, a KNN-based stream classifier that introduces inference-time adaptation for jointly addressing concept drift and  class imbalance. Unlike existing approaches that adapt primarily by modifying stored instances, \ours{} maintains a simple sliding-window memory and adapts predictions through dynamic class-prior adjustment. 
Our experimental evaluation demonstrated that \ours{} achieves competitive performance across a wide range of imbalance and drift scenarios, while performing particularly well under dynamic class imbalance and concept drift. 
These results indicate that effective adaptation in KNN-based stream classification does not require continuous expensive memory operations.

Instead, shifting adaptation to inference offers a simple and responsive mechanism for handling rapidly changing stream conditions.

More broadly, our findings suggest that adaptation in memory-based stream classifiers need not rely exclusively on increasingly sophisticated modifications to stored instances. 
The simplicity of \ours{} allows it to react quickly in highly dynamic settings without permanently altering the stored evidence. Actively managed memories, by contrast, can improve information retention over longer horizons when the assumptions underlying their management mechanisms hold, but may prematurely modify or discard instances based on uncertain evidence of change. Memory-based and inference-time adaptations should therefore be viewed as complementary design choices, with memory adaptations offering benefits for information preservation over long horizons and inference adaptations better equipped to tackle dynamic imbalance ratio shifts. 

Future work will extend inference-time adaptation beyond class-prior adjustment by exploring feature-aware drift signals, and dynamic neighborhood selection, as well as its integration with memory-based adaptation.



\begin{credits}

\subsubsection{\discintname}
The authors have no competing interests to declare that are
relevant to the content of this article.
\end{credits}
%
%
%
\bibliographystyle{splncs04}
\bibliography{stream_knn_references}
\end{document}